\documentclass{article}
\usepackage{arxiv}
\usepackage{subcaption} 
\usepackage[utf8]{inputenc} 
\usepackage[T1]{fontenc}    
\usepackage{hyperref}       
\usepackage{url}            
\usepackage{booktabs}       
\usepackage{amsfonts}       
\usepackage{nicefrac}       
\usepackage{ulem}

\usepackage{float} 
\usepackage{enumitem}
\usepackage{microtype}      
\usepackage{lipsum}
\usepackage{graphicx}
\graphicspath{ {./images/} }
\usepackage{natbib}
\usepackage{amsmath}
\usepackage{comment}
\usepackage[monochrome]{xcolor}
\newcommand{\DRS}[2][blue]{{\textcolor{#1}{#2}}}

\title{Multimodal Neural Operators for Real-Time Biomechanical Modelling of Traumatic Brain Injury}

\author{
 Anusha Agarwal\\
  Thomas Jefferson High School for Science and Technology\\
  Alexandria, 6560 Braddock Rd\\
  VA 22312, United States\\
  \texttt{2026aagarwal@tjhsst.edu} \\
  \\
  Johns Hopkins Whiting School of Engineering\\
  Baltimore, 3400 N Charles St\\
  MD 21218, United States\\
  \texttt{aagarw57@jh.edu} \\
   \And
 Dibakar Roy Sarkar \\
  Johns Hopkins Whiting School of Engineering\\
  Baltimore, 3400 N Charles St\\
  MD 21218, United States\\
  \texttt{droysar1@jh.edu} \\
  \And
 Somdatta Goswami\thanks{Corresponding author: \texttt{somdatta@jhu.edu}} \\
  Johns Hopkins Whiting School of Engineering\\
  Baltimore, 3400 N Charles St\\
  MD 21218, United States\\
  \texttt{somdatta@jhu.edu} \\
}

\begin{document}
\maketitle
\begin{abstract}

\textbf{Background:} Traumatic brain injury (TBI) remains a major public health concern, with over 69 million cases annually worldwide. \DRS{Accurate patient-specific biomechanical modeling is critical for injury risk assessment, but it requires integrating heterogeneous data sources such as volumetric neuroimaging, scalar demographic parameters, and acquisition metadata. Conventional finite element solvers can perform this modeling, yet they remain too computationally expensive for time-sensitive clinical settings. Neural operators have emerged as a promising alternative by learning resolution-invariant mappings between function spaces at orders-of-magnitude faster inference. However, while recent work has introduced parameter-conditioned and multi-input operator architectures for incorporating auxiliary scalar or geometric inputs, the systematic integration of volumetric medical imaging with heterogeneous scalar metadata within an operator learning framework remains underexplored, particularly for biomechanical prediction tasks involving patient-specific anatomical and demographic variability.}

\textbf{Objective:} This study \DRS{presents a systematic investigation of multimodal neural operator architectures for brain biomechanics, evaluating strategies for fusing} heterogeneous input modalities \DRS{such as volumetric anatomical imaging, scalar demographic features, and acquisition parameters} to predict full-field brain displacement fields \DRS{from MRE data}.

\textbf{Methods:} We reformulated TBI modeling as a multimodal operator learning problem and proposed two fusion strategies: field projection for Fourier Neural Operator (FNO) based architectures (broadcasting scalars onto spatial grids) and branch decomposition for Deep Operator Networks (DeepONet) (separate encoding with multiplicative fusion). Four architectures (FNO, Factorized FNO (F-FNO), Multi-Grid FNO (MG-FNO), DeepONet) were extended with multimodal fusion mechanisms and evaluated on 249 \textit{in vivo} Magnetic Resonance Elastography (MRE) datasets across physiologically relevant frequencies (20 to 90 Hz).

\textbf{Results:} \DRS{DeepONet achieved the highest accuracy on real displacement fields (MSE = 0.0039, 90.0\% accuracy) with the fastest inference (3.83 it/s) and fewest parameters (2.09M), while MG-FNO achieved the best performance on imaginary fields (MSE = 0.0058, 88.3\% accuracy) with the lowest GPU memory among FNO variants (7.12 GB). No single architecture dominated across all criteria, revealing distinct trade-offs between accuracy, spatial fidelity, and computational cost.}

\textbf{Conclusion:}  \DRS{The results demonstrate that neural operators augmented with multimodal fusion mechanisms can accurately predict full-field brain displacement from heterogeneous biomedical inputs, with inference times orders of magnitude faster than finite element solvers. The systematic comparison of fusion strategies and architectures provides practical guidance for selecting operator learning approaches in biomedical settings where heterogeneous data integration is required.}
\end{abstract}

\textbf{Keywords:} Multimodal neural operators, Heterogeneous data fusion, Fourier Neural Operator (FNO), Deep Operator Network (DeepONet), Traumatic brain injury, Magnetic Resonance Elastography (MRE), Patient-specific modeling, Real-time prediction, Digital twins

\section{Introduction}
\label{sec:intro}

Neural operators have revolutionized scientific computing by learning mappings between infinite-dimensional function spaces, enabling orders-of-magnitude speedups over traditional numerical solvers for partial differential equations \cite{kovachki2023neural,azizzadenesheli2024neural}. Architectures such as the Fourier Neural Operator (FNO) \cite{li2021fourier} and Deep Operator Network (DeepONet) \cite{lu2019deeponet} have demonstrated remarkable success across diverse domains, from fluid dynamics and weather forecasting to materials modeling and climate prediction \cite{pathak2022fourcastnet,wen2023real,goswami2022deep}.  \DRS{However, most neural operator formulations assume relatively homogeneous inputs, typically scalar or vector fields defined on structured grids. Several recent advances have begun to relax this assumption. Wen et al. \cite{wen2022u} demonstrated that scalar parameters can be broadcast as constant fields onto spatial grids to augment FNO inputs for multiphase flow prediction. Jin et al. \cite{jin2022mionet} introduced MIONet, which extends DeepONet to multiple input functions defined on different domains via tensor product composition, with theoretical guarantees for multi-input operator approximation. Subsequent work has combined these ideas with Fourier layers \cite{jiang2024fourier}, geometry-aware conditioning \cite{peyvan2025fusion,li2023geometry}, and temporal conditioning mechanisms for hybrid solver-operator frameworks \cite{oommen2024rethinking}. Vectorized Conditional Neural Fields further condition predictions on PDE parameters for parametric generalization \cite{hagnberger2024vectorized}. While these methods have successfully incorporated auxiliary scalar or geometric inputs in engineering applications, they have not been evaluated in settings that require fusing high-dimensional volumetric medical imaging with heterogeneous patient-specific metadata for biomechanical prediction. Biomedical operator learning poses distinct challenges, including inter-subject anatomical variability, mixed continuous and categorical covariates (e.g., age, sex, scan direction), and spatially varying tissue properties, which motivate a dedicated investigation.}

Consider traumatic brain injury (TBI), affecting nearly 69 million people annually worldwide and causing over 214,000 hospitalizations and 200 deaths per day in the United States alone \cite{cdc_tbi_data}. Predicting patient-specific brain tissue deformation under mechanical loading, which is critical for clinical triage and injury risk assessment, requires integrating multiple heterogeneous data sources: three-dimensional anatomical MRI capturing individual brain morphology, scalar viscoelastic tissue properties from magnetic resonance elastography (MRE), demographic features such as age and sex, acquisition parameters including loading frequency and scan direction, and binary anatomical masks constraining the solution domain \cite{matney2022scope}.

Finite element (FE) models have emerged as the gold standard for characterizing brain biomechanics. When combined with high-resolution neuroimaging, FE simulations can predict subject-specific brain responses under controlled loading \cite{hu2022advanced,bahreinizad2023mri,gomez2019acceleration}. However, these models are computationally demanding, often requiring hours of runtime per subject, limiting their integration into time-sensitive clinical settings \cite{griffiths2022finite}. Recent machine learning approaches have attempted to accelerate TBI modeling \cite{wu2022realtime}, but conventional neural networks trained on discretized FE simulations lack resolution invariance and fail to generalize across input discretizations or unseen parameter combinations.

Neural operators (NOs) \DRS{provide a promising alternative by learning resolution-invariant mappings between input and output function spaces. Although recent work has demonstrated parameter-conditioned and geometry-aware operator learning in engineering domains \cite{wen2022u,jin2022mionet,peyvan2025fusion,li2023geometry}, these methods have not been systematically evaluated for biomechanical prediction from clinical neuroimaging data, where the input space combines volumetric anatomical images with mixed-type patient metadata.}

\DRS{In this work, we adapt and extend established neural operator architectures to the problem of predicting full-field brain displacement under harmonic loading from multimodal \textit{in vivo} data. Specifically, we integrate structural MRI-derived anatomy with scalar demographic and acquisition parameters within operator learning frameworks, applying two fusion strategies drawn from the broader literature: field projection (broadcasting scalars onto spatial grids, following the approach of \cite{wen2022u}) for FNO-based architectures, and branch decomposition (separate encoding with multiplicative fusion, related to the multi-branch paradigm of \cite{jin2022mionet}) for DeepONet. Using a dataset of 249 MRE experiments across physiologically relevant frequencies (20--90~Hz) \cite{bayly2021mr}, we compare four neural operator architectures, namely FNO, Factorized FNO (F-FNO), Multi-Grid FNO (MG-FNO), and DeepONet, for their ability to predict three-dimensional brain displacement fields. Our goals are threefold: (1)~to evaluate how existing multimodal fusion strategies perform when applied to volumetric neuroimaging data combined with heterogeneous patient metadata for biomechanical prediction, (2)~to systematically compare the accuracy, computational efficiency, and failure modes of four neural operator architectures on \textit{in vivo} brain MRE data, and (3)~to identify architectural trade-offs relevant to potential deployment in time-sensitive biomedical
settings.}

The remainder of this paper is organized as follows: Section 2 describes the dataset. Section 3 presents the multimodal operator learning framework. Section 4 reports quantitative and qualitative evaluations. Section 5 concludes with future directions for multimodal operator learning.

\section{Dataset}
\label{sec:dataset}

The present study leverages recently developed, spatially resolved neuroimaging datasets that capture the biomechanics of the living human brain under controlled loading conditions \cite{bayly2021mr}. These datasets provide three-dimensional, time-resolved displacement and strain fields, along with complementary anatomical and microstructural information, to enable subject-specific modeling and evaluation of computational traumatic brain injury (TBI) models. Data acquisition, as reported in \cite{bayly2021mr}, was conducted across three research sites - Washington University in St. Louis (WUSTL), the University of Delaware (UD) and the National Institutes of Health (NIH), with protocols tailored to different aspects of brain biomechanics.

\subsection{Dataset Composition}

The complete dataset includes multimodal imaging data from 300 participants (approximately 100 per site), encompassing a broad age range (14-80 years) and balanced sex representation. Participants were healthy volunteers recruited to capture normative biomechanical variability across age and sex. Each subject underwent one of three primary biomechanical imaging protocols, depending on site:
\begin{itemize}[leftmargin=*]
\item {Tagged MRI (NIH):} Full-brain, time-resolved deformation fields were captured during impulsive, sub-injury rotational loading, with 18 ms temporal resolution and comprehensive brain coverage. These data provide a direct measure of dynamic tissue displacement under controlled acceleration, enabling validation of computational models.
\item {Magnetic Resonance Elastography (WUSTL):} Brain displacement and strain fields were measured in response to harmonic skull motion across multiple excitation frequencies (20-90 Hz) with 3 mm isotropic voxel resolution. This modality captures frequency-dependent dynamic behavior and wave propagation patterns within brain tissue.
\item {High-Resolution MRE (UD):} Spatially resolved maps of brain tissue viscoelastic properties-including complex shear modulus, stiffness, and damping ratio-were acquired at multiple frequencies (30-70 Hz) with 1.5 mm isotropic voxel resolution. These high-resolution data allow for subject-specific material property assignment in computational models.
\end{itemize}

All participants additionally underwent anatomical (T1- and T2-weighted) and diffusion-weighted MRI to provide structural context. These scans enable tissue segmentation, white matter tractography, and vascular feature extraction. Participant demographics and anthropometrics-including age, sex, height, weight, and brain volume-are also included, facilitating stratified analyses and investigation of age- and sex-specific effects on brain biomechanics. Acquisition parameters, such as scan direction and vibration frequency, are provided: direction (anterior-posterior [AP] or left-right [LR]) indicates scan orientation, while vibration frequency (20-90 Hz) corresponds to the mechanical excitations used to generate MRE displacement fields.

\textcolor{blue} {It is important to note that, although the broader dataset includes multiple imaging modalities and acquisition sites, the present study only uses the WUSTL dataset for MRE displacement prediction. This source provides comprehensive measurements of brain dynamics under mechanical loading. In this dataset, 52 subjects were studied across various vibration frequencies, leading to a final dataset size of 249 samples. This restriction limits the diversity of the training distribution and should be considered when interpreting generalization and clinical applicability.}

\subsection{Data Characteristics}

To ensure cross-site consistency, all acquisitions are performed on Siemens 3T scanners with site-specific head coils, and harmonization procedures are applied to minimize inter-scanner variability. Imaging protocols include:
\begin{itemize}[leftmargin=*,nosep]
\item Tagged MRI displacement and strain fields derived using the harmonic phase finite element (HARP-FE) method.  
\item MRE displacement data preprocessed with rigid-body motion estimation and phase unwrapping to recover full-field displacement vectors.  
\item High-resolution MRE displacement data inverted via nonlinear finite element methods to estimate storage and loss moduli ($G'$, $G''$), stiffness ($\mu$), and damping ratio ($\xi$).  
\end{itemize}

Derived outputs include full 3D displacement vectors, strain tensors, frequency-dependent material property maps, and summary scalar metrics such as maximum principal strain (MPS) and octahedral shear strain (OSS). Anatomical and diffusion-weighted images are processed through skull-stripping, atlas-based segmentation, and diffusion tensor imaging (DTI) tractography to generate label maps and axonal orientation fields, supporting integration of structural features into computational models.

\begin{figure}[h]
\centering
\includegraphics[width=0.6\textwidth]{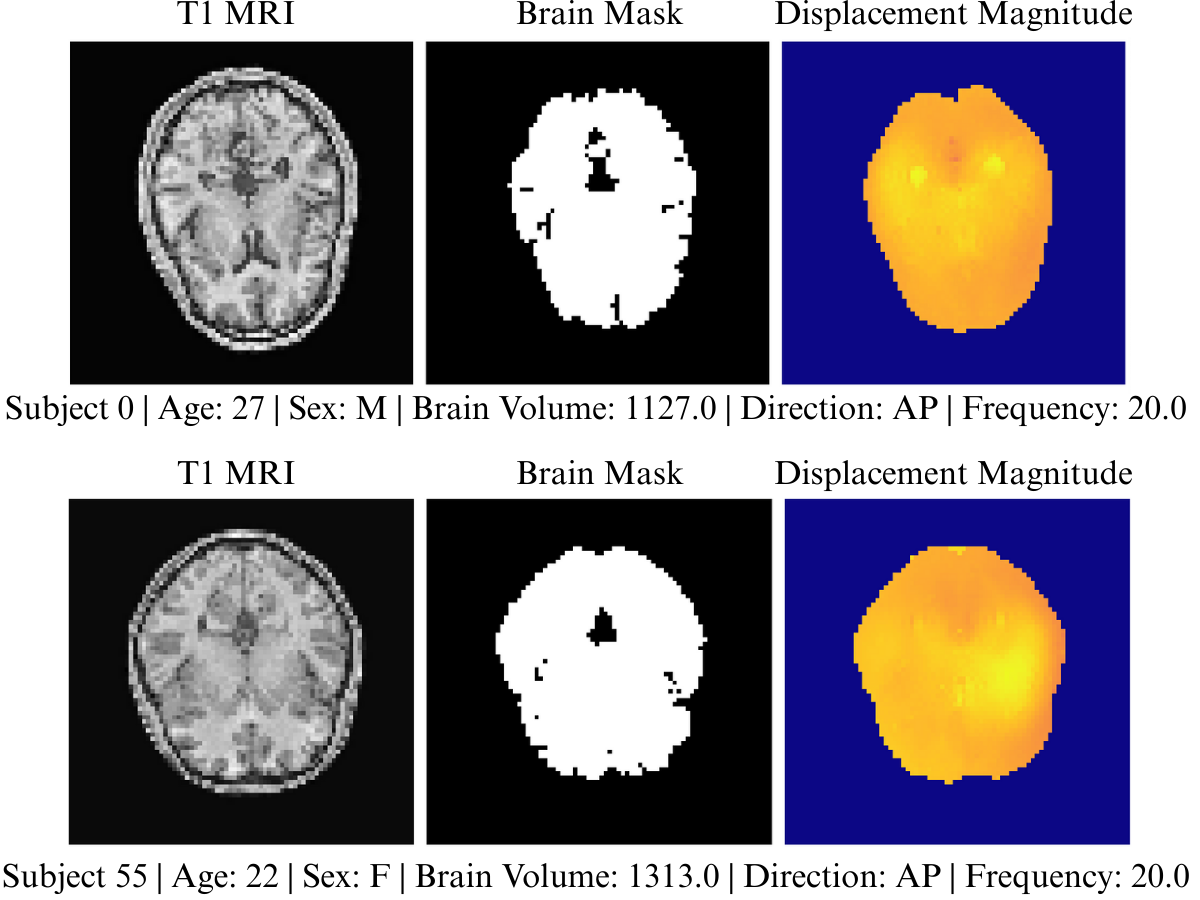}
\caption{Representative visualization of the dataset, showing T1-weighted anatomical MRI (left), brain mask (middle), and displacement fields (right), along the central axial cross section for two samples. Displacement magnitude combines both the real and imaginary components of the displacement field. Direction relates to how the T1-weighted MRI was scanned and frequency relates to the rate at which mechanical vibrations are applied for MRE.}
\label{fig:sample_data}
\end{figure}

Representative examples of the dataset, including the attributes utilized in this study, are provided in Figure~\ref{fig:sample_data}. 

\subsection{Data Access and Format}

All datasets are publicly disseminated through the Brain Biomechanics Imaging Resources repository hosted on the Neuroimaging Tools and Resources Collaboratory (NITRC) at \url{https://www.nitrc.org/projects/bbir}. Shared data include raw and processed displacement fields, strain fields, viscoelastic property maps, head kinematics, and corresponding anatomical MRI. Data are provided in Neuroimaging Informatics Technology Initiative (NIfTI) format. Anatomical data are registered to the MNI-152 atlas space and resampled at 0.8 mm isotropic resolution to facilitate cross-subject comparisons. All datasets are de-identified, collected under IRB-approved protocols, and shared under the GNU GPL v3.0 license.

\subsection{Data Processing}

All data were stored in HDF5 format.  \DRS{The dataset comprises 52 unique subjects totaling 249 samples. We partition the data using a subject-level split to prevent data leakage: 36 subjects (171 samples) for training, 5 subjects (25 samples) for validation, and 11 subjects (53 samples) for testing. This ensures that no subject appears in more than one partition, so the model is always evaluated on entirely unseen anatomies.} We developed a custom PyTorch-compatible data loader to parse each HDF5 file, extracting both imaging and metadata into structured dictionaries. After parsing and preprocessing each sample, all relevant tensors and metadata were stored in a list of Python dictionaries, with each dictionary representing a complete training sample. This structure allows flexible access during batching, data augmentation, and multimodal model construction. Each dictionary contains all fields (T1 volume, displacement field, brain mask, and metadata) for a single subject. These were cached in memory for rapid batching and I/O-efficient training.

\subsubsection{Data Scaling} To ensure numerical stability and accelerate convergence during training, we applied a series of normalization steps to both imaging and demographic data. Scaling was performed globally across the entire dataset where applicable, using min-max normalization schemes tailored to each data modality. 

T1-weighted MRI volumes were normalized on a per-voxel basis using global minimum and maximum intensity values computed across all subjects. This strategy preserves spatial contrast patterns between individuals while bringing voxel values into the $[0, 1]$ range. For each voxel coordinate $(x, y, z)$, we computed:
\[
\text{voxel}_{\text{min}}(x, y, z) = \min_{i} \text{T1}_i(x, y, z), \quad
\text{voxel}_{\text{max}}(x, y, z) = \max_{i} \text{T1}_i(x, y, z)
\]
where $i$ indexes subjects. The final T1 volumes were scaled as:
\[
\text{T1}_{\text{scaled}} = \frac{\text{T1} - \text{voxel}_{\text{min}}}{\text{voxel}_{\text{max}} - \text{voxel}_{\text{min}}}
\]
Displacement fields were globally scaled using the minimum and maximum of all displacement magnitudes across the dataset. This preserves relative deformation strength while ensuring that all values lie in the unit interval. Scalar demographic features (age and brain volume) were normalized using dataset-wide minimum and maximum values. Age was linearly scaled to $[0, 1]$ using standard min-max scaling:
\[
\text{Age}_{\text{scaled}} = \frac{\text{Age} - \text{Age}_{\min}}{\text{Age}_{\max} - \text{Age}_{\min}}
\]
Brain volume, on the other hand, was scaled to the range $[0.1, 0.9]$ to preserve margin for potential out-of-distribution values. Invalid or missing demographic entries (e.g., age $\leq 0$) were encoded with a sentinel value of $-1$ and excluded from model input where appropriate.

\paragraph{Brain Masking.}
To restrict the analysis to anatomically relevant regions, we applied a binary mask that delineates brain tissue from the surrounding background. The mask assigns a value of 1 to voxels corresponding to brain tissue and 0 to non-brain regions, thereby excluding non-brain voxels from subsequent computation. Importantly, the mask does not remove noise directly; rather, it isolates the spatial extent of the brain so that preprocessing and model training are performed only within this region. Masking was applied prior to scaling to prevent artificial amplification of background values and to reduce unnecessary computation outside the brain. This ensures that normalization and intensity scaling are applied exclusively to meaningful brain voxels, improving training stability and reconstruction quality. It is important to note that the brain masks are provided in the dataset, and we utilize these pre-existing masks to constrain the learning process to anatomically relevant regions.

\section{Neural Operators}

Neural operators (NOs) are a class of deep learning architectures that learn mappings between infinite-dimensional function spaces, offering a powerful alternative to traditional physics-based solvers. Rather than approximating pointwise functions, NOs learn the solution operator of a physical system, mapping input fields such as material properties or boundary conditions directly to output fields such as displacements. Critically, NOs are resolution-invariant and generalize across discretizations, enabling efficient surrogate modeling of complex PDE systems. They have shown state-of-the-art performance in fluid dynamics, elasticity, and other domains where long-range spatial dependencies and high-dimensional fields are essential.

The first NO architecture, the Deep Operator Network (DeepONet), was introduced in 2019 \cite{lu2019deeponet}, building on the universal approximation theorem for operators by Chen \& Chen \cite{chen1995operator}. DeepONet employs a dual-network structure: a branch network encodes input functions sampled at sensor points, while a trunk network encodes spatial or spatio-temporal output coordinates. Their inner-product fusion enables accurate and efficient surrogate modeling, and DeepONet has since been applied to a range of high-dimensional scientific problems.

Another major class of NOs uses spectral integral formulations. The Fourier Neural Operator (FNO) parameterizes convolution kernels in the Fourier domain, capturing global structures efficiently and achieving orders-of-magnitude faster inference than conventional solvers \cite{li2021fourier}. Building on this, the Factorized FNO (F-FNO) introduces separable spectral layers and improved residual connections, reducing parameter counts while allowing deeper networks \cite{ffno}. The Multi-Grid Tensorized FNO (MG-FNO) further decomposes the input domain hierarchically and tensorizes Fourier kernels, enabling high-resolution simulations with fewer weights \cite{kossaifi2023multi}. These architectures provide complementary advantages, improved representation power, parameter efficiency, and scalability, making them well-suited for capturing heterogeneous, high-resolution brain deformations.

NOs have been successfully applied across diverse domains, including materials modeling \cite{goswami2022deep, ROYSARKAR2026118926}, earthquake response prediction \cite{goswami2025neural}, weather forecasting \cite{kurth2023fourcastnet}, fluid dynamics \cite{goswami2023physics}, and biomedical imaging \cite{maier2022known}. They have enabled breakthroughs in solving differential equations \cite{cao2024laplace}, accelerated real-time prediction of complex dynamics \cite{kontolati2024learning, karumuri2025physics}, and coupled seamlessly with numerical solvers for hybrid approaches \cite{wang2025time}. Applications also include inverse problems and parameter estimation in heterogeneous systems \cite{sarkar2025adaptive, sarkar2025interface}, time-dependent PDE modeling via physics-informed temporal operators \cite{nayak2025ti, mandl2025physics}, and resolution-independent formulations that maintain accuracy across discretizations \cite{bahmani2025resolution}. Recent advances include separable physics-informed architectures that break the curse of dimensionality \cite{mandl2025separable}, synergistic multitask frameworks for efficient PDE solution \cite{kumar2025synergistic}, and stochastic modeling of structural systems under natural hazards \cite{goswami2025neural, thiagarajan2025accelerating}.

\begin{figure}
  \centering
  \includegraphics[width=\textwidth]{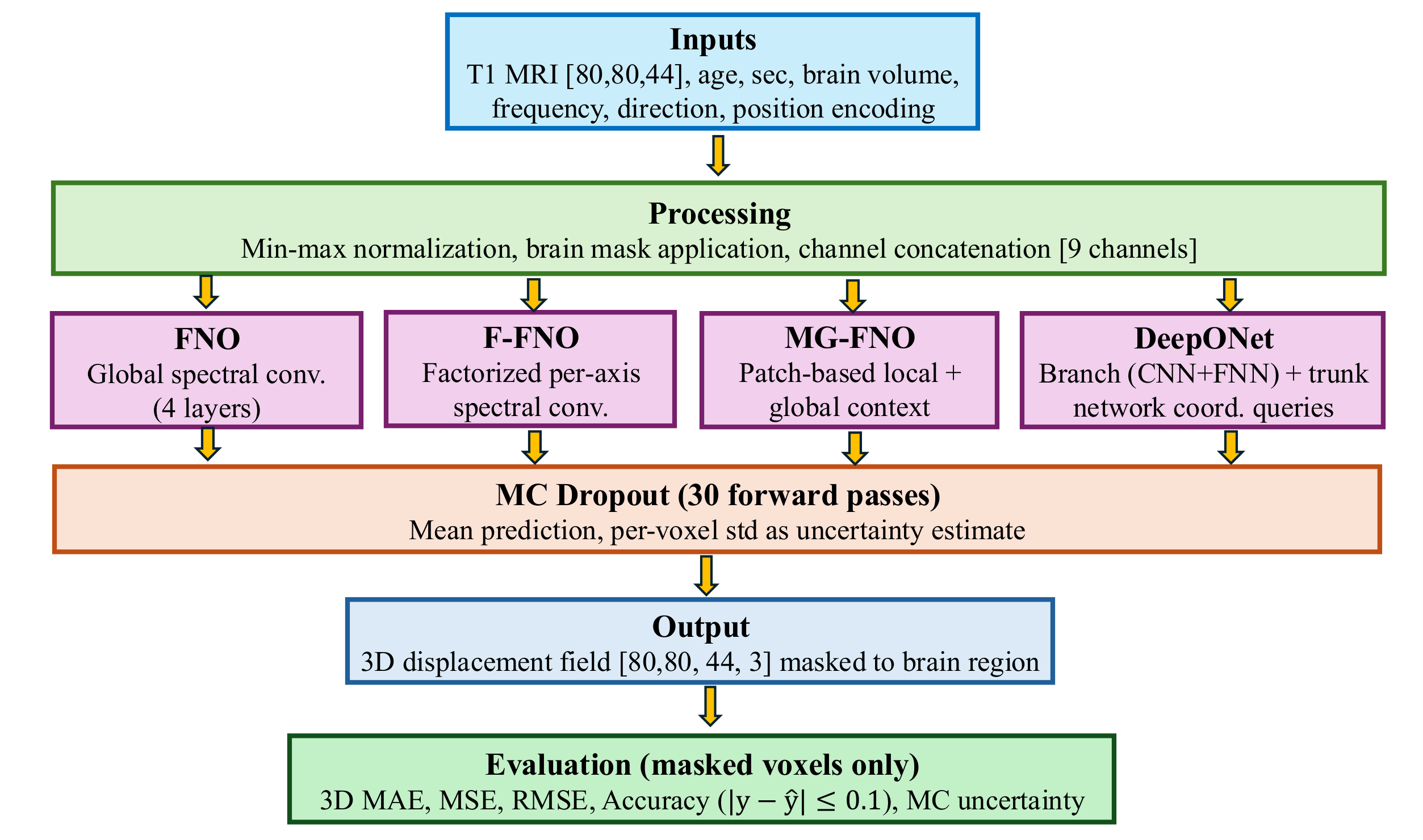}
  \caption{\DRS{General framework for neural operator-based brain displacement prediction. All architectures share a common input pipeline: T1-weighted MRI volumes ($80 \times 80 \times 44$) are concatenated with five scalar subject features and 3D positional encodings to form a 9-channel input tensor, normalized to $[0,1]$ using training-set statistics and masked to brain regions. The input is processed by one of four neural operator architectures, each differing in how spatial information is handled. Monte Carlo Dropout (30 forward passes) provides per-voxel uncertainty estimates alongside the predicted 3D displacement field ($u_x, u_y, u_z$). Performance is evaluated on masked voxels using MAE, MSE, RMSE, and tolerance-based accuracy ($\delta = 0.1$).}}
  \label{fig:operatorflow}
\end{figure}

In this work, we develop multimodal extensions of four state-of-the-art neural operator architectures for patient-specific TBI modeling: FNO \cite{li2021fourier}, F-FNO \cite{ffno}, MG-FNO \cite{kossaifi2023multi}, and DeepONet \cite{lu2019deeponet}. Each architecture is augmented with multimodal fusion mechanisms to handle heterogeneous inputs combining anatomical imaging, scalar demographic parameters, and geometric constraints. All models were implemented in PyTorch \cite{paszke2019pytorch}, and we detail their multimodal architectural designs, fusion strategies, training procedures, and hyperparameter optimization in the following sections. \textcolor{blue}{Figure \ref{fig:operatorflow} outlines the general framework for neural operator-based brain displacement prediction. The four neural operator architectures evaluated in this work share a common input pipeline but differ substantially in how they process spatial information. The standard FNO applies a joint 3D spectral convolution over the full $80 \times 80 \times 44$ volume, giving it strong long-range dependency modeling but limiting its ability to resolve localized displacement patterns due to spectral mode truncation. The Factorized FNO (F-FNO) reduces this computational burden by decomposing the 3D spectral operation into three separate 1D transforms along each spatial axis, lowering parameter count at the cost of missing cross-axis frequency interactions. The Multi-Grid FNO (MG-FNO) partitions the volume into local patches of size $20 \times 20 \times 22$ voxels, each processed independently by the FNO operator but augmented with a downsampled global T1 context channel, allowing simultaneous resolution of fine-grained local structure and global anatomical context. DeepONet departs most significantly from the FNO class by separating the operator into a branch network consisting of a CNN processing the T1 image combined with scalar subject features, and a trunk network queried at arbitrary $(x, y, z)$ coordinate locations, making it resolution independent but introducing a larger train-validation gap due to its coordinate-based formulation. All four architectures are trained with an identical masked MSE loss, subject-level data splits, and MC Dropout uncertainty estimation, enabling a controlled comparison of their architectural prediction biases for brain displacement.}

\DRS{We focus exclusively on neural operator architectures in this study because they offer resolution invariance and parameter generalization properties that alternative surrogate approaches lack. CNN-based models (e.g., 3D U-Net) are tied to fixed discretizations and do not generalize across resolutions \cite{kovachki2023neural}, while physics-informed neural networks require explicit PDE formulations that are unavailable for the \textit{in vivo} MRE setting considered here and are typically trained per-instance rather than across multiple subjects and conditions.}

\subsection{Fourier Neural Operator} 

\paragraph{Data Representation.} After T1 and scalar data have been preprocessed, scalar data must be projected onto a 3D grid to allow the model to access all scalar features at every single voxel in the T1 image grid. Age, brain volume, sex, frequency, and direction are the 5 selected scalar features for input. These features are then projected onto a field of the same depth, height, and width as the input of the T1 image so that each feature is in shape [1, W, H, D]. From here, all features are stacked along the first input channel, creating an input of size [6, W, H, D]. A key design decision was to project scalar metadata into dense 3D fields instead of concatenating them at later layers in the network. While both approaches are valid, projection at the input stage enforces spatial conditioning, ensuring that metadata influences the model uniformly across the entire volume from the very beginning of training. This allows the network to learn interactions between global metadata and local structural features in the T1 image, rather than deferring this integration until later layers, where spatial correspondence might already be abstracted away. In practice, this strategy encourages the model to treat scalar features as fundamental contextual priors, directly modulating voxel-level representations.

\paragraph{Positional Encoding.}
As FNO learns in a 3D spatial domain, adding positional features enhances the model’s ability to reason in a location-informed grid rather than relying solely on appearance features. To encode spatial position, we generate continuous positional grids across the x, y, and z dimensions using the PyTorch \texttt{linspace} function. Each grid spans the normalized range [0, 1], where 0 represents one boundary of the axis and 1 represents the opposite boundary. This normalization ensures scale invariance, making the encoding independent of the absolute image dimensions and instead tied to relative voxel location. Each positional grid is appended along the first channel of the existing input, allowing the model to distinguish between structurally similar but spatially distinct features (e.g., left vs. right hemispheres). With this addition, the final input for FNO prediction takes the shape [9, W, H, D], consisting of the original six scalar/T1 channels and three positional channels.

\subsubsection{FNO Architecture} 
We employ the Fourier Neural Operator (FNO), which leverages spectral convolution layers to capture long-range spatial dependencies. Below we outline the specific architecture used in our study, which operates as follows:

\textbf{Lifting Layer:} \\
  The input function \( u(x) \), discretized on a uniform grid, is first lifted to a higher-dimensional representation using a pointwise (fully connected) linear layer:
  \[
  v_0(x) = P(u(x)), \quad \text{where } P : \mathbb{R}^d \to \mathbb{R}^m
  \]
  This increases the feature dimension from the input space to a latent space suitable for learning complex operators.

\textbf{Fourier Layers:} \\
  Each Fourier layer performs a global operation through the following steps:
  \begin{itemize}
    \item \textit{Fourier Transform:} Convert the spatial representation \( v_j(x) \) into the frequency domain using the fast Fourier transform (FFT):
    \[
    \hat{v}_j(k) = \mathcal{F}[v_j](k)
    \]
    
    \item \textit{Spectral Convolution:} Apply a learned complex-valued linear transformation in the frequency domain:
    \[
    \hat{v}_{j+1}(k) = W(k) \hat{v}_j(k)
    \]
    Typically, only the lowest \( K \) Fourier modes are retained and updated, reducing computational cost and acting as a form of regularization.
    
    \item \textit{Inverse Fourier Transform:} Transform the result back to the spatial domain:
    \[
    v_{j+1}(x) = \mathcal{F}^{-1}[\hat{v}_{j+1}](x)
    \]
    
    \item \textit{Nonlinearity:} Apply a non-linear activation function (in this case, GELU \cite{hendrycks2016gelu}) pointwise in the spatial domain.
  \end{itemize}
  We repeat this process is repeated in four layers, enabling the model to learn complex nonlinear mappings in a globally coupled manner.

\begin{figure}
  \centering
  \includegraphics[width=15cm]{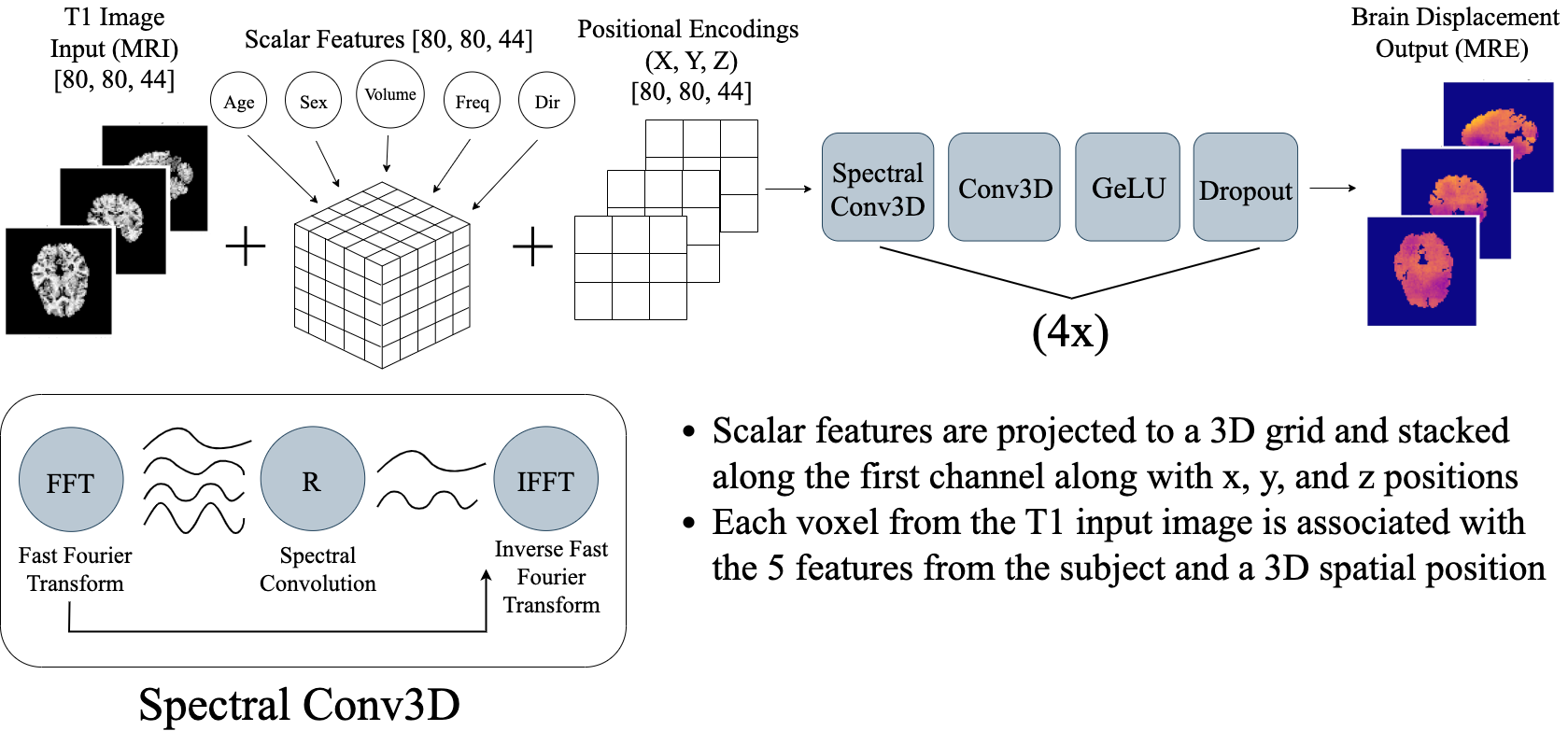}
  \caption{Fourier Neural Operator (FNO) Architecture for Brain Displacement Prediction. The network takes T1 MRI images [80, 80, 44] as primary input, which are augmented with scalar subject features (age, sex, volume, frequency, direction) and 3D positional encodings (X, Y, Z coordinates), all projected to matching grid dimensions [80, 80, 44]. These multi-modal inputs are concatenated along the channel dimension and processed through the core Spectral Conv3D operation, which performs convolution in the frequency domain via Fast Fourier Transform (FFT), spectral convolution (R), and Inverse FFT. The spectral convolution is followed by standard 3D convolution, GeLU activation, and dropout layers (repeated 4 times) to predict 3D brain displacement fields for Magnetic Resonance Elastography (MRE).}
  \label{fig:fnoarch}
\end{figure}

\textbf{Projection Layer:} \\
  After the sequence of Fourier layers, a final pointwise linear layer maps the output from the latent space back to the desired output function space:
  \[
  u_{\text{out}}(x) = Q(v_n(x)), \quad \text{where } Q : \mathbb{R}^m \to \mathbb{R}^d
  \]
The schematic of this architecture is presented in Figure~\ref{fig:fnoarch}.

\subsubsection{Factorized Fourier Neural Operator (F-FNO).} 
We build upon the Factorized Fourier Neural Operator (F-FNO) architecture proposed by Tran et. al \cite{ffno}. This model enhances the traditional FNO framework by factorizing the spectral convolution into one-dimensional operations along each spatial axis, reducing the number of learnable parameters and the computational cost by over two orders of magnitude while retaining accuracy. 

Given an input tensor $u \in \mathbb{R}^{\mathbb{R}^{B \times C_{\text{in}} \times D \times H \times W}}$, we first apply a lifting layer 
\begin{equation}
v_0 = \sigma\!\left( \text{Conv}_{1\times 1}(u) \right), \quad v_0 \in \mathbb{R}^{B \times d_{\text{model}} \times D \times H \times W},
\end{equation}
where $\sigma$ denotes the GELU activation and $d_{\text{model}} = 64$ is the latent channel dimension.  

Each subsequent F-FNO layer applies three factorized spectral convolutions, one per spatial dimension, followed by pointwise mixing and a residual connection:
\begin{equation}
v_{\ell+1} = v_\ell + \sigma\!\left( W_2 \, \sigma\!\left( W_1 \, \mathcal{F}^{-1}\Big( \sum_{j=1}^3 \mathcal{P}_j\big( \mathcal{F}_j(v_\ell) \cdot \Theta_j \big) \Big) \right)\right),
\end{equation}
where $\mathcal{F}_j$ and $\mathcal{F}^{-1}$ denote the forward and inverse one-dimensional Fourier transforms along dimension $j \in \{D,H,W\}$, $\Theta_j$ are complex-valued learnable weights truncated to the lowest $m_j$ modes, and $\mathcal{P}_j$ projects the truncated spectrum back to the original resolution via zero-padding. The $1 \times 1 \times 1$ convolutions $W_1$ and $W_2$ serve as channel mixers, and $\sigma$ is again the GELU nonlinearity. In our setting, we retain $m_1=40$, $m_2=40$, and $m_3=12$ modes respectively.  

Stacking $L=4$ such layers yields the encoded representation $v_L$, which is then mapped to the three-dimensional displacement field via a projection head:
\begin{equation}
\hat{u} = \text{Conv}_{1\times 1}\!\left(\sigma\!\left(\text{Conv}_{1\times 1}(v_L)\right)\right), \quad \hat{u} \in \mathbb{R}^{B \times 3 \times D \times H \times W}.
\end{equation}

This architecture preserves the global receptive field of the Fourier operator while substantially lowering memory usage and training cost. By truncating to low-frequency modes, the model also enforces smoothness and reduces aliasing artifacts, which is particularly beneficial for large-scale 3D scientific data.

\subsubsection{Multi-Grid Fourier Neural Operator (MG-FNO).}  
\begin{figure}
  \centering
  \includegraphics[width=15cm]{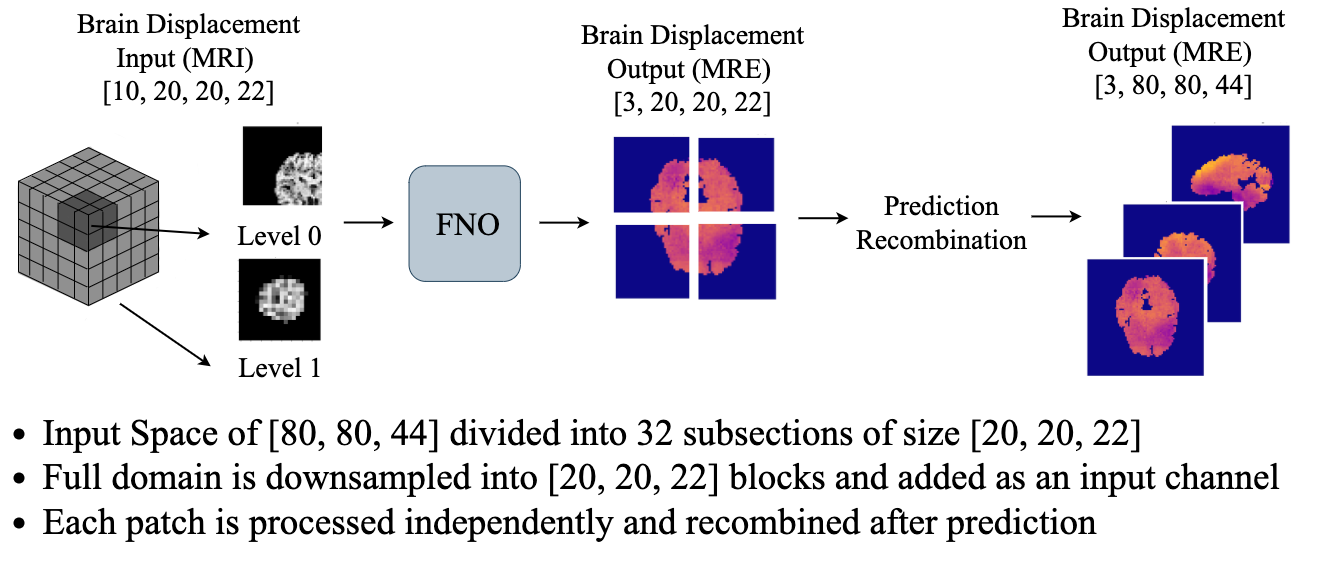}
  \caption{Multi-Grid Fourier Neural Operator (FNO) Architecture for Hierarchical Brain Displacement Prediction. The input domain [80, 80, 44] is spatially decomposed into 32 non-overlapping subsections of size [20, 20, 22] to enable multi-scale processing. The brain MRI data is processed at multiple resolution levels: Level 0 operates on downsampled blocks [20, 20, 22], while Level 1 processes the full domain context. Each level is fed into separate FNO networks that independently predict 3-component displacement fields [3, 20, 20, 22] for their respective scales. The multi-scale predictions are then recombined through a prediction recombination module to reconstruct the final high-resolution brain displacement output [3, 80, 80, 44] for MRE analysis. This hierarchical approach allows the model to capture both local fine-grained deformation patterns within individual patches and global long-range dependencies across the entire brain domain, improving computational efficiency while maintaining prediction accuracy for biomechanical modeling.}
  \label{fig:mgarch}
\end{figure}

To further improve scalability, spatial continuity, and localization, we integrated a multigrid approach into the FNO structure, inspired by the framework proposed by Kossaifi, et. al \cite{kossaifi2023multi}. Rather than processing the entire 3D domain holistically, the MG-FNO partitions the input volume into smaller, non-overlapping subdomains (patches). Each patch is processed independently, enabling the network to learn localized deformation patterns with greater efficiency while still incorporating essential global context. In our implementation, the domain is divided into patches of size $[20, 20, 22]$, which evenly partition the T1-weighted MRI volumes of shape $[80, 80, 44]$. For each local patch, we additionally provide a downsampled global representation of the T1 input. This global context is generated by interpolating the full-resolution T1 channel to a coarser scale (factor of 4), and then resizing it back to match the patch dimensions. By concatenating this global T1 context with each local patch, the network maintains awareness of long-range dependencies and overall brain geometry, while still learning highly localized features. \DRS{It is important to note that cross-patch communication is not explicitly enforced during inference; global information is incorporated implicitly through the shared downsampled context concatenated to each patch, rather than through direct interactions between neighboring patches.} The resulting patch-level input has an augmented channel dimension $[10, 20, 20, 22]$, compared to the baseline FNO input $[9, 20, 20, 22]$. Once all patches are processed, predictions are reassembled into the full displacement field by directly mapping each patch back into its corresponding subdomain location. This patch-based reconstruction ensures consistency across the full domain while enabling fine-grained learning at the patch level. The architecture is visualized in Figure~\ref{fig:mgarch}.

\subsection{DeepONet Architecture}

The Deep Operator Network (DeepONet) is a neural architecture designed to learn nonlinear operators that map between function spaces, making it particularly effective for modeling mappings from input functions (e.g., images) to output functions (e.g., deformation fields) \cite{lu2019deeponet}. 

DeepONet consists of two primary components: a \textit{branch network} and a \textit{trunk network}. The branch network encodes the input functions and associated scalar features, while the trunk network encodes the spatial coordinates where the output is evaluated. The final output is obtained via an inner-product fusion of branch and trunk features, enabling the model to decouple spatial and functional dependencies.

\paragraph{Branch Network.}  
In our implementation, the branch network is composed of multiple subnetworks, each processing a specific modality or scalar feature:  

\begin{itemize}
    \item \textbf{CNN branch:} Processes the 2D slices of the T1-weighted MRI to extract high-level image features. The CNN consists of three convolutional layers with batch normalization and pooling (two max-pooling and one average pooling), followed by fully connected layers with dropout to produce a 300-dimensional embedding.
    \item \textbf{FNN branches:} Each scalar attribute, scan direction, vibration frequency, sex, brain volume, and age, is processed through a separate fully connected feedforward network (FNN), each producing a 300-dimensional embedding. These subnetworks capture the effect of acquisition parameters and participant-specific demographics on the predicted displacement.
\end{itemize}

The outputs of all branch subnetworks are combined multiplicatively to form a single 300-dimensional branch representation that encodes both anatomical and scalar information.

\paragraph{Trunk Network.}  
The trunk network encodes the spatial coordinates of the voxels at which the displacement is to be predicted. In our implementation, the trunk network is a multilayer perceptron with linear layers and ReLU activations, mapping the 3D coordinates \( (x, y, z) \) to a feature vector of the same dimensionality as the branch output (300 dimensions).  

\paragraph{Output Fusion.}  
The final displacement prediction at each voxel is computed via an elementwise inner product between the branch embedding and the corresponding trunk embedding for each output dimension:
\[
\mathbf{u}(x) \approx \sum_{i=1}^{p} b_i(u) \cdot t_i(x), \quad \mathbf{u}(x) = (u_x, u_y, u_z),
\]
where \(b_i(u)\) is the \(i\)-th branch feature and \(t_i(x)\) is the \(i\)-th trunk feature. In our implementation, the branch output is split into three 100-dimensional vectors corresponding to the \(x\), \(y\), and \(z\) displacement components, which are fused with the corresponding trunk features via the inner product.  

\begin{figure}
  \centering
  \includegraphics[width=14cm]{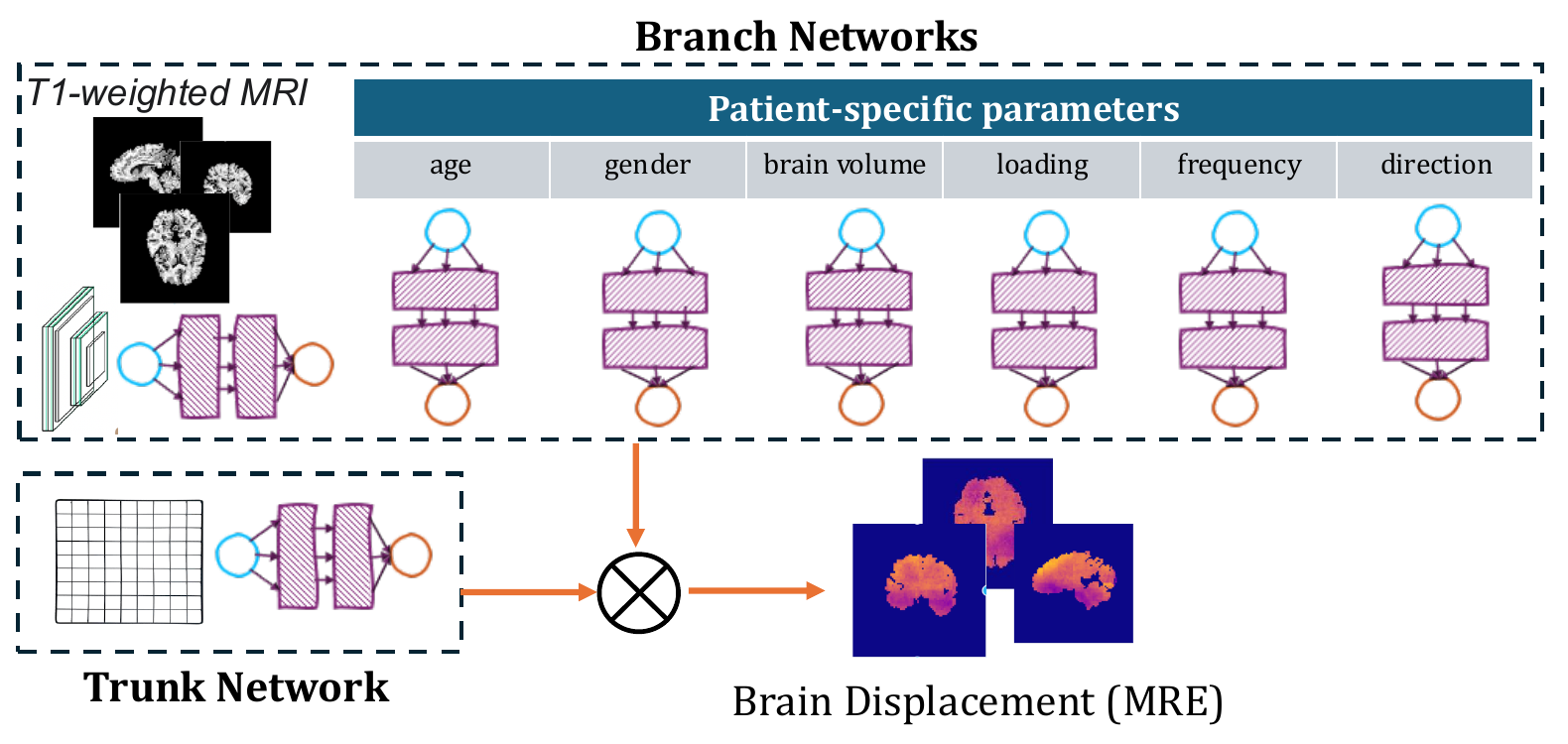}
    \caption{Deep Operator Network (DeepONet) Architecture for Brain Displacement Prediction. The architecture employs six branch networks and one trunk network to map multimodal inputs to 3D displacement fields. A CNN branch processes T1-weighted MRI slices [80$\times$80$\times$44] through convolutional layers with batch normalization and pooling to extract 300D anatomical features, while five separate FNN branches encode scalar parameters (scan direction, vibration frequency, sex, brain volume, age) into 300D embeddings each. All branch outputs are fused via element-wise multiplication to create a unified 300D representation encoding both anatomical and demographic information. The trunk network processes 3D spatial coordinates ($x$, $y$, $z$) through multilayer perceptrons to generate 300D spatial basis functions at each voxel location. For displacement prediction, the 300D branch vector is partitioned into three 100D segments corresponding to $x$-, $y$-, and $z$-components, which undergo inner-product fusion with corresponding trunk segments via Einstein summation. This operator learning framework generates continuous 3D displacement fields [Batch $\times$ N\_voxels $\times$ 3] by decoupling spatial and functional dependencies, enabling prediction of brain tissue deformation from multimodal anatomical and acquisition parameters for MR elastography applications.}
  \label{fig:donarch}
\end{figure}

The multimodal nature of our DeepONet architecture is realized through the branch network decomposition strategy. Each input modality, including anatomical imaging via CNN branch and scalar features (scan direction, vibration frequency, sex, brain volume, age) via individual FNN branches, is independently encoded into a 300-dimensional latent representation by its dedicated subnetwork. These modality embeddings are fused via elementwise multiplication to create a unified multimodal representation capturing the joint influence of all inputs on brain displacement. This multiplicative fusion allows the network to learn nonlinear interactions between heterogeneous data types, enabling each modality to modulate others. The resulting combined branch representation encodes spatial anatomical features and demographic acquisition parameters, ensuring predictions are conditioned on complete multimodal subject context. Architecture visualization is in Figure~\ref{fig:donarch}.

\subsection{Model Training and Validation}
The models were trained to predict the three displacement components as outputs, using the T1-weighted image and associated scalar features as inputs. Prior to training, displacement fields were scaled and masked to enforce stability, reduce numerical variance, and ensure that predictions were constrained to anatomically valid regions. This setup allowed the networks to directly learn the physical continuity and spatial structure of displacement without relying on intermediate transformations or surrogate targets.

All models were trained on the Rockfish cluster at Johns Hopkins University, specifically on the \texttt{ica100} partition. Each node in this partition is equipped with 64 CPU cores across 2 sockets, 256 GB of RAM, and 4 NVIDIA A100 GPUs, providing the computational power necessary for training large-scale operator learning models.

\begin{figure}
  \centering
  \includegraphics[width=\textwidth]{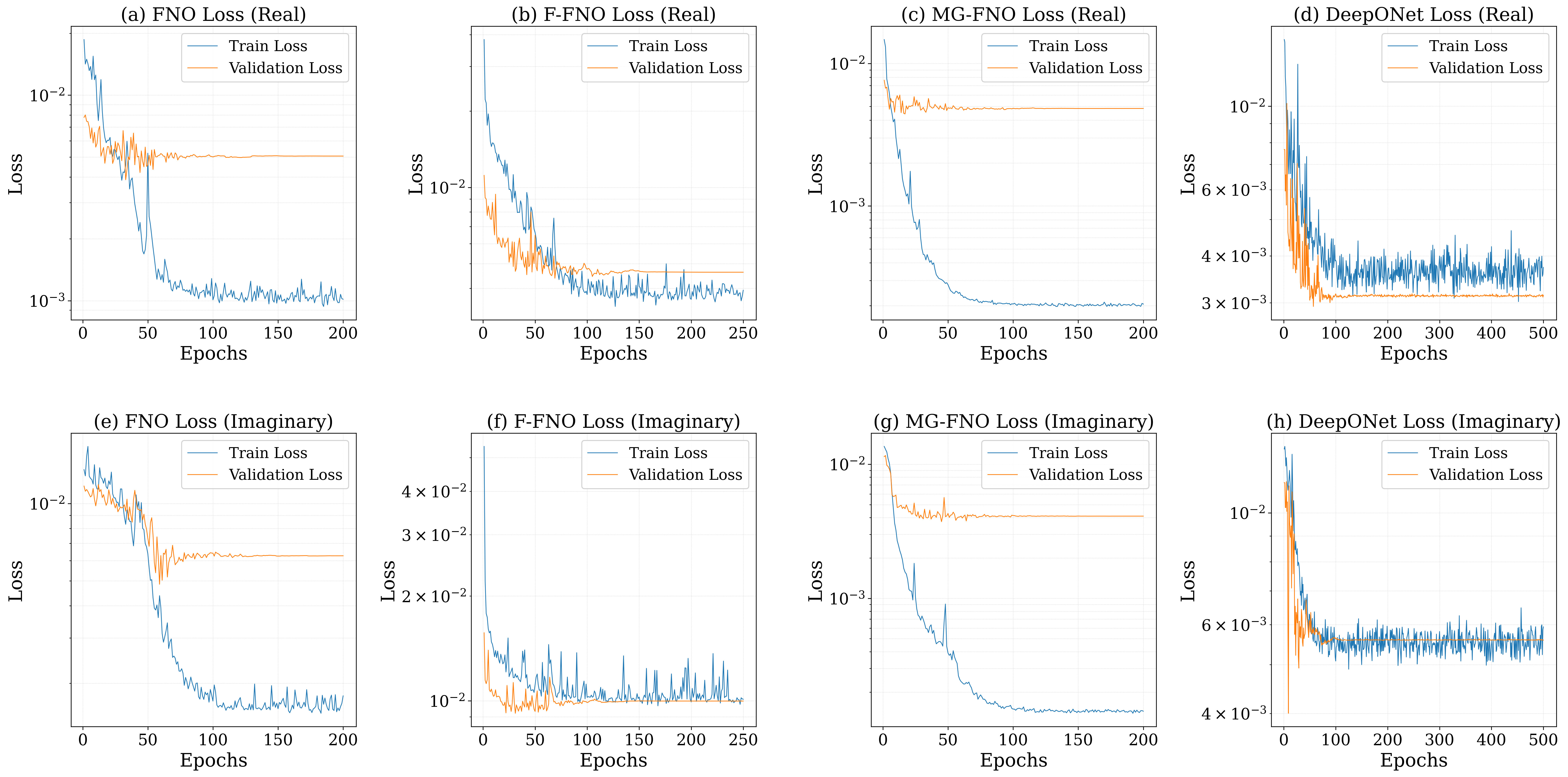}
  \caption{\DRS{Training and validation losses across all architectures for real (top) and imaginary (bottom) displacement fields. The FNO and F-FNO exhibit a notable gap between training and validation loss, particularly on the real component, reflecting overparameterization relative to the limited training set. The MG-FNO shows noisier convergence due to patch-level loss aggregation. DeepONet converges with the smallest train-validation gap, consistent with its compact architecture. All models converge faster on the imaginary component, which exhibits smoother spatial structure.}}
  \label{fig:losses}
\end{figure}

\DRS{Training schedules were tailored to each architecture based on convergence behavior: the FNO and MG-FNO were trained for 200 epochs, the F-FNO for 250 epochs, and the DeepONet for 500 epochs. The loss history for each training run is reported in Figure~\ref{fig:losses}, showing both training and validation loss progression. A persistent gap between training and validation loss is visible across most architectures, particularly for the FNO and F-FNO on the real component. This divergence is likely attributable to two factors: first, the validation set comprises only 25 samples drawn from approximately 5 unique subjects, which may not adequately represent the variability of the full data distribution; second, the high-capacity FNO-based architectures (up to 1.42B parameters) are substantially overparameterized relative to the training set size (174 samples), making them prone to memorizing subject-specific patterns that do not transfer to the limited validation cohort. The MG-FNO exhibits noisier loss curves due to its patch-based processing, where loss is aggregated over independently processed subdomains. DeepONet shows the most stable convergence with the smallest train-validation gap, consistent with its compact parameterization (2.09M).
}

\subsubsection{Hyperparameter Tuning}  
A learning rate of $1\times 10^{-3}$ was used throughout most of the training process, with the AdamW optimizer \cite{loshchilov2017adamw} (\texttt{weight\_decay} = $1\times 10^{-5}$) to balance fast convergence with regularization. To further stabilize optimization, we employed a \texttt{ReduceLROnPlateau} scheduler, which adaptively decayed the learning rate by a factor of 0.5 if validation loss did not improve for 10 epochs.  

Loss was computed using a masking voxel-wise Mean Squared Error (MSE) criterion such that gradients only accumulated within anatomically valid brain regions. Specifically, losses were first calculated per voxel and then multiplied by the binary brain mask before reduction, ensuring that boundary regions or background voxels did not bias the optimization process.  

During each epoch, both training and validation losses were logged. The best-performing checkpoint (based on validation loss) was automatically saved to disk, allowing recovery of the optimal model state for inference. 

\paragraph{Modes Selection.} 
In Fourier Neural Operator (FNO) training, the primary tunable parameter is the number of Fourier modes retained in each spatial dimension $(x, y, z)$. For an input domain with size $N_x \times N_y \times N_z$, the maximum number of independent modes in each dimension is given by:
\[
k_\text{max}^x = \frac{N_x}{2} + 1, \quad 
k_\text{max}^y = \frac{N_y}{2} + 1, \quad 
k_\text{max}^z = \frac{N_z}{2} + 1.
\]
Retaining fewer modes acts as a low-pass filter and helps prevent overfitting while reducing computational cost.

For both the standard FNO and the fully-factorized FNO (F-FNO), we found that using approximately half of the maximum available modes in each dimension provided the best performance, resulting in $[k_x, k_y, k_z] = [40, 40, 12]$. For the multi-grid FNO (MG-FNO), it was advantageous to use the full set of available modes, corresponding to $[k_x, k_y, k_z] = [20, 20, 12]$, to better capture high-frequency components conditioned on global inputs.

\section{Results}
We evaluated the performance of the benchmarked neural operator models using the \textit{mean squared error} (MSE) metric, defined as  

\begin{equation}
\text{MSE} = \frac{1}{N} \sum_{i=1}^{N} \left( y_i - \hat{y}_i \right)^2,
\end{equation}
 
where $N$ is the number of samples, $y_i$ represents the ground-truth displacement or field value, and $\hat{y}_i$ denotes the corresponding prediction from the model. 

\DRS{MSE was chosen as the primary performance indicator because it penalizes large deviations more strongly than other error metrics, making it particularly sensitive to the sharp displacement gradients present in traumatic brain injury (TBI) simulations. For completeness, we also report mean absolute error (MAE), root mean squared error (RMSE), and accuracy.}  

\DRS{Accuracy is defined as the fraction of valid (masked) voxels for which the predicted displacement falls within an absolute tolerance of the ground truth value:
\begin{equation}
    \text{Accuracy} = \frac{1}{V} \sum_{i=1}^{V} 
    \mathbf{1}\left[ \left| y_i - \hat{y}_i \right| \leq \delta \right],
\end{equation}
where $V$ is the total number of valid brain-masked voxels across all test samples, $y_i$ and $\hat{y}_i$ denote the ground-truth and predicted displacement values respectively, $\mathbf{1}[\cdot]$ is the indicator function, and $\delta = 0.1$ is the absolute tolerance threshold. Since all displacement fields are normalized to $[0, 1]$ via global min-max scaling prior to evaluation, $\delta = 0.1$ corresponds to a 10\% tolerance relative to the full normalized displacement range.}

Table~\ref{tab:direction_comparison} reports the predictive performance of the models on real and imaginary displacement fields, while Table~\ref{tab:compute} summarizes computational efficiency in terms of iterations per second. \DRS{In this study, we train separate models for the real and imaginary components of the MRE displacement field. This separation reflects the underlying physics of the measurement: in MRE, tissue motion under harmonic excitation is represented as a complex-valued displacement field, where the real component corresponds to the in-phase elastic response and the imaginary component captures the out-of-phase viscous response of brain tissue \cite{manduca2021mr}. Both components are required for downstream inversion algorithms that estimate the complex shear modulus $G^* = G' + iG''$, where the storage modulus $G'$ is derived primarily from the real part and the loss modulus $G''$ from the imaginary part \cite{mcgarry2015suitability}. Clinically relevant injury metrics such as MPS are computed from spatial gradients of the full displacement field, making accurate prediction of both components a necessary prerequisite. Errors in either component propagate directly into strain estimates and material property reconstructions.}

\begin{table}[h]
  \centering
  \DRS{\caption{Performance Comparison of Benchmarked Architectures on Real and Imaginary Displacement Fields. MSE is highlighted as the primary evaluation metric.}
  \small
  \begin{tabular}{llllll}
    \toprule
    \textbf{Direction} & \textbf{Model} & \textbf{MAE} & \textbf{MSE} & \textbf{RMSE} & \textbf{Accuracy} \\
    \midrule
    Real 
        & FNO & $0.0438 \pm 0.0335$ & $0.0051 \pm 0.0068$ &  $0.0567 \pm 0.0439$ & $0.8813 \pm 0.1582$ \\
        & F-FNO          & $0.0427 \pm 0.0329$ & $0.0053 \pm 0.0069$ & $0.0572 \pm 0.0451$ & $0.8853 \pm 0.1481$ \\
        & MG-FNO          & $0.0397 \pm 0.0315$ & $0.0044 \pm 0.0061$ & $0.0515 \pm 0.0417$ & $0.8938 \pm 0.1481$ \\
        & DeepONet                & $\mathbf{0.0350 \pm 0.0314}$ & $\mathbf{0.0039 \pm 0.0057}$ & $\mathbf{0.0463 \pm 0.0421}$ & $\mathbf{0.9000 \pm 0.1465}$ \\
    \midrule
    Imaginary 
        & FNO & $0.0468 \pm 0.0472$ & $0.0074 \pm 0.0153$ & $0.0608 \pm 0.0606$ & $0.8628 \pm 0.1832$ \\
        & F-FNO          & $0.0543 \pm 0.0460$ & $0.0090 \pm 0.0187$ & $0.0720 \pm 0.0620$ & $0.8471 \pm 0.1703$ \\
        & MG-FNO          & $\mathbf{0.0410 \pm 0.0446}$ & $\mathbf{0.0058 \pm 0.0126}$ & $\mathbf{0.0523 \pm 0.0554}$ & $\mathbf{0.8825 \pm 0.1866}$ \\
        & DeepONet                & $0.0475 \pm 0.0390$     & $0.0064 \pm 0.0106$     & $0.0612 \pm 0.0512$     & $0.8669 \pm 0.1739$ \\
    \bottomrule
  \end{tabular}
  \label{tab:direction_comparison}}
\end{table}

\begin{table}[h]
  \centering
  \DRS{\caption{Compute Efficiency of Benchmarked Architectures. Higher iterations per second indicate faster inference.}
  \small
  \begin{tabular}{lrrrr}
    \toprule
    \textbf{Model} & \textbf{Iter/sec} $\uparrow$ & \textbf{Parameters} $\downarrow$
    & \textbf{Train time (min)} $\downarrow$
    & \textbf{GPU mem (GB)} $\downarrow$ \\
    \midrule
    FNO      & 0.65          & 1.42B              & \textbf{57.1}  & 40.71 \\
    F-FNO    & 0.52          & 6.95M              & 244.7          & 49.98 \\
    MG-FNO   & 0.08          & 353.95M            & 180.5          & 7.12  \\
    DeepONet & \textbf{3.83} & \textbf{2.09M}     & 60.9           & \textbf{4.11} \\
    \bottomrule
  \end{tabular}
  \label{tab:compute}}
\end{table}

In the following subsections, we present results for each neural operator individually. Within each subsection, we discuss the performance of the vanilla operator and its variants in detail, highlighting comparative accuracy, error metrics, and architectural differences. A full comparison of each model's predictions and errors is presented in Table~\ref{tab:direction_comparison}.

\subsection{FNO and Variant Architectures Evaluation}

\begin{figure}
  \centering
  \includegraphics[width=\textwidth]{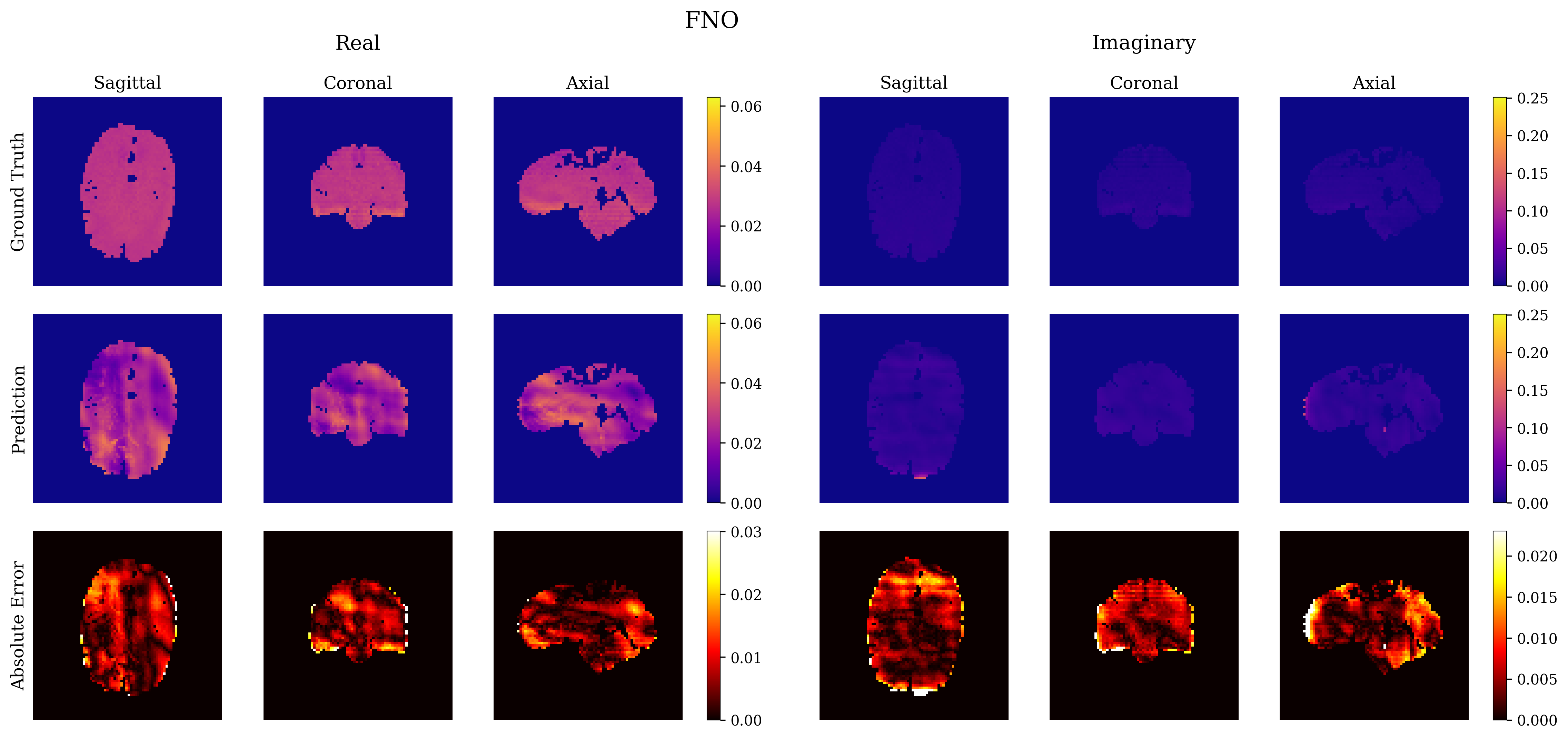}
  \caption{\DRS{FNO predictions and absolute error maps across sagittal, coronal, and axial planes for real (left) and imaginary (right) displacement fields. The model captures global displacement structure in both components but exhibits higher errors near cortical boundaries and high-gradient regions. For the imaginary component, errors are concentrated at localized boundary regions.}}
  \label{fig:fno_prediction}
\end{figure}

The baseline FNO achieved an MSE of \DRS{$0.0051 \pm 0.0068$} on real displacement fields, with corresponding MAE of \DRS{$0.0438 \pm 0.0335$} and RMSE of \DRS{$0.0567 \pm 0.0439$}. \DRS{The prominent error for FNO is found in the boundaries as shown in a representative sample in Fig.~\ref{fig:fno_prediction}}. In the imaginary sample, there is a sharp error in very localized at specific part of the brain. Notably, these errors are concentrated near regions of sharp spatial variation, indicating that the model struggles specifically in areas where accurate gradient resolution is most critical. Rather than capturing these sharp transitions, the FNO smooths and redistributes them along coordinate directions, leading to the observed axis-aligned patterns. Performance declined on imaginary displacement fields, where the MSE rose substantially to \DRS{$0.0074 \pm 0.0153$}, highlighting the difficulty of capturing abrupt transitions in magnitude. Computationally, the FNO processed approximately \DRS{0.65} iterations per second, representing moderate efficiency relative to its variants. These observations motivate the use of architectural modifications, such as factorization and global conditioning, to enhance the model’s ability to capture both high-frequency local features and long-range dependencies.

\begin{figure}
  \centering
  \includegraphics[width=\textwidth]{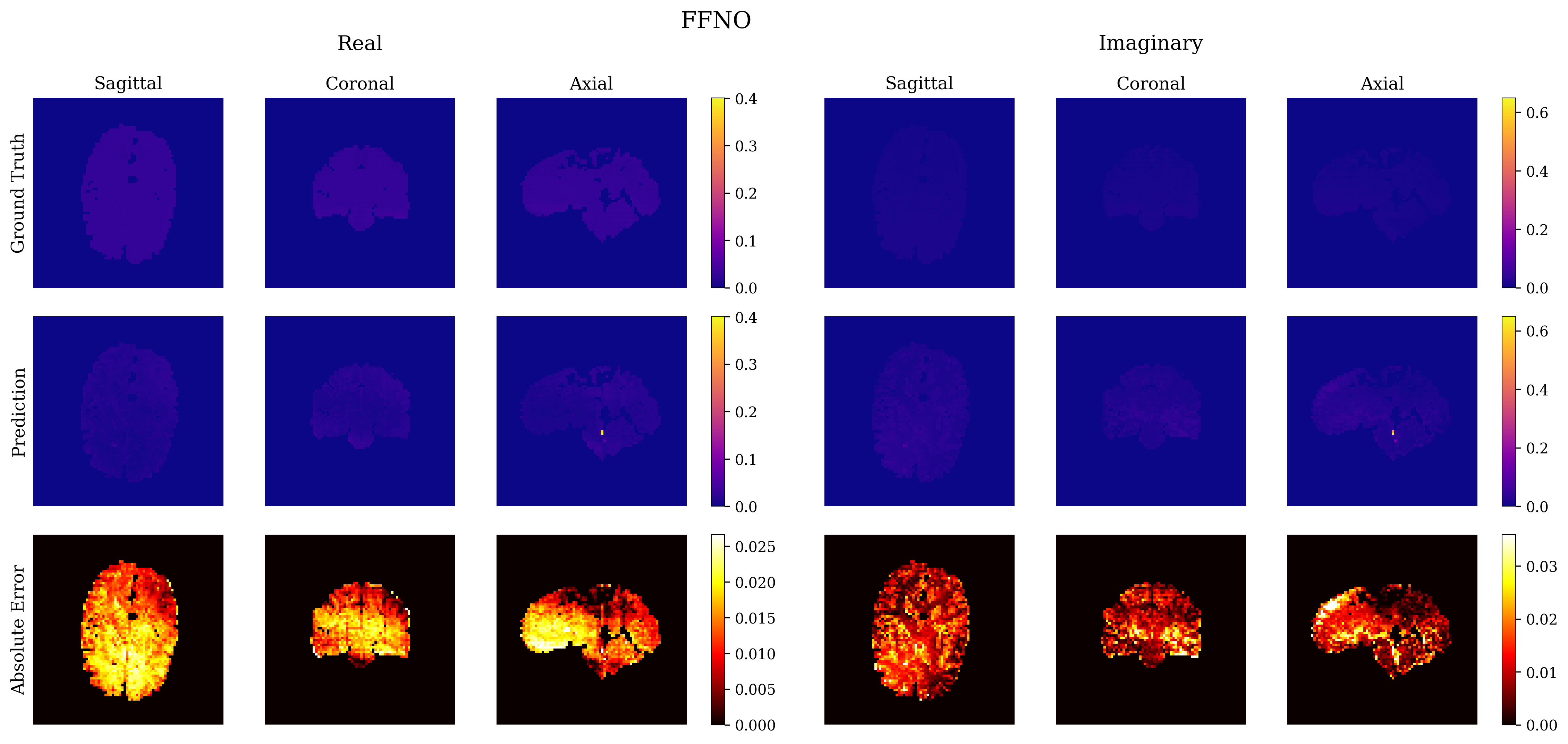}
  \caption{\DRS{F-FNO predictions and absolute error maps across sagittal,coronal, and axial planes for real (left) and imaginary (right) displacement fields. The model produces spatially smoother error distributions compared to the baseline FNO, but exhibits isolated single-voxel outliers with disproportionately high predicted magnitudes (visible in axial predictions for both components), suggesting occasional instability in the factorized Fourier representation. Outside these outliers, errors remain moderate and distributed across the brain volume.}}
  \label{fig:ffno_prediction}
\end{figure}

\DRS{\paragraph{Factorized Fourier Neural Operator.} The F-FNO did not improve predictive accuracy over the baseline FNO on real displacement fields, achieving an MSE of $0.0053 \pm 0.0069$ compared to $0.0051 \pm 0.0068$ for the baseline, with comparable accuracy ($0.8853 \pm 0.1481$ vs.\ $0.8813 \pm 0.1582$). Performance on imaginary fields was also weaker, with an MSE of $0.0090 \pm 0.0187$ and the lowest accuracy among all architectures ($0.8471 \pm 0.1703$), suggesting that the low-rank factorization of Fourier weight matrices sacrifices representational capacity needed to capture high-frequency displacement variations. While factorization reduces the parameter count by approximately $204\times$ ($6.95$M vs.\ $1.42$B), the F-FNO was designed for large-scale problems where full spectral convolutions become a bottleneck \cite{ffno}. In our setting, the relatively small input volume ($80 \times 80 \times 44$) means full spectral convolutions are already efficient, so the factorization overhead dominates rather than providing a speedup, resulting in slower inference ($0.52$ it/s vs.\ $0.65$ it/s) and longer training ($244.7$ min vs.\ $57.1$ min). Qualitatively, the F-FNO produced smoother predictions and reduced the axis-aligned streaking artifacts of the baseline, but exhibited isolated single-voxel outliers with anomalously high values (Figure~\ref{fig:ffno_prediction}), indicating occasional instability in the factorized representation. These findings suggest that factorization is better suited to higher-dimensional or larger-resolution operator learning problems than the moderate-scale volumetric setting considered here.}

\begin{figure}
  \centering
  \includegraphics[width=\textwidth]{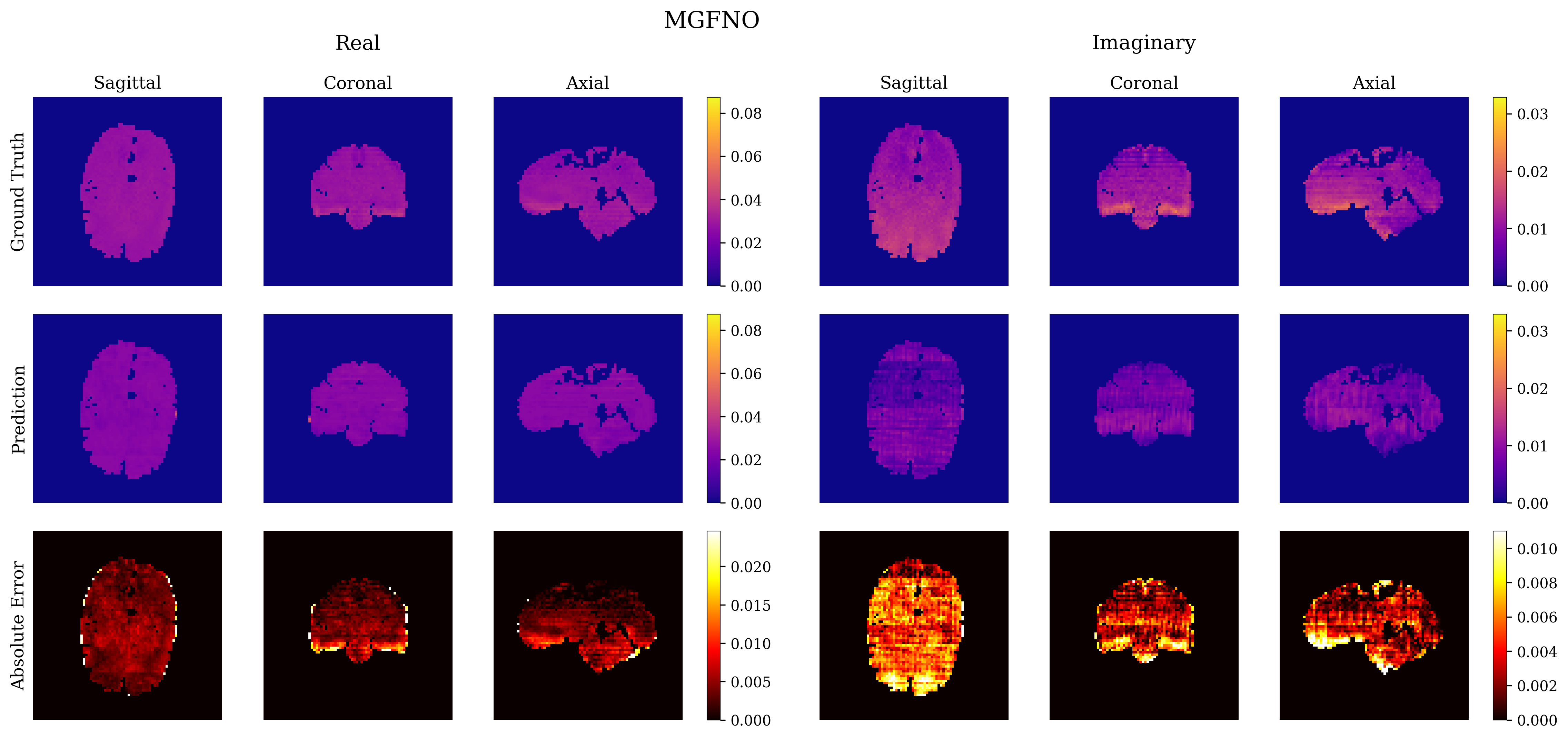}
  \caption{\DRS{MG-FNO predictions and absolute error maps across sagittal, coronal, and axial planes for real (left) and imaginary (right) displacement fields. The model achieves the closest visual agreement with ground truth among all architectures, with notably lower error magnitudes. However, the imaginary predictions exhibit horizontal banding patterns visible in the coronal and axial views, a residual artifact of the patch-based processing where independent patch predictions introduce subtle discontinuities at patch boundaries.}}
  \label{fig:mgfno_prediction}
\end{figure}

\DRS{\paragraph{Multi-Grid Fourier Neural Operator.} The MG-FNO achieved the strongest performance on imaginary displacement fields, with an MSE of $0.0058 \pm 0.0126$ and accuracy of $0.8825 \pm 0.1866$, outperforming all other architectures on this component. On real displacement fields, it ranked second with an MSE of $0.0044 \pm 0.0061$ and accuracy of $0.8938 \pm 0.1481$, behind DeepONet ($0.9000 \pm 0.1465$). Relative to the baseline FNO, the MG-FNO reduced MSE by approximately $13.7\%$ on the real component and $21.6\%$ on the imaginary component. This improvement can be attributed to the multigrid architecture, which captures hierarchical frequency content and facilitates information exchange between coarse- and fine-resolution levels. This structure mitigates the tendency of standard FNOs to over-smooth sharp transitions, allowing the model to better resolve localized features such as abrupt variations in displacement magnitude (Figure~\ref{fig:mgfno_prediction}). However, the imaginary predictions exhibit visible horizontal banding artifacts arising from the patch-based processing, where independent patch predictions introduce subtle discontinuities at patch boundaries. From a computational standpoint, the MG-FNO was the slowest architecture tested ($0.08$ it/s), reflecting the overhead of processing multiple patches sequentially, despite having a moderate parameter count ($353.95$M) and the lowest GPU memory footprint among FNO-based models ($7.12$ GB). These results suggest that the multigrid strategy offers an effective accuracy-memory trade-off at the cost of inference speed, and that addressing patch boundary artifacts through overlapping patches or boundary blending could
further improve spatial continuity.}

\subsection{DeepONet Evaluation} 

\begin{figure}
  \centering
  \includegraphics[width=\textwidth]{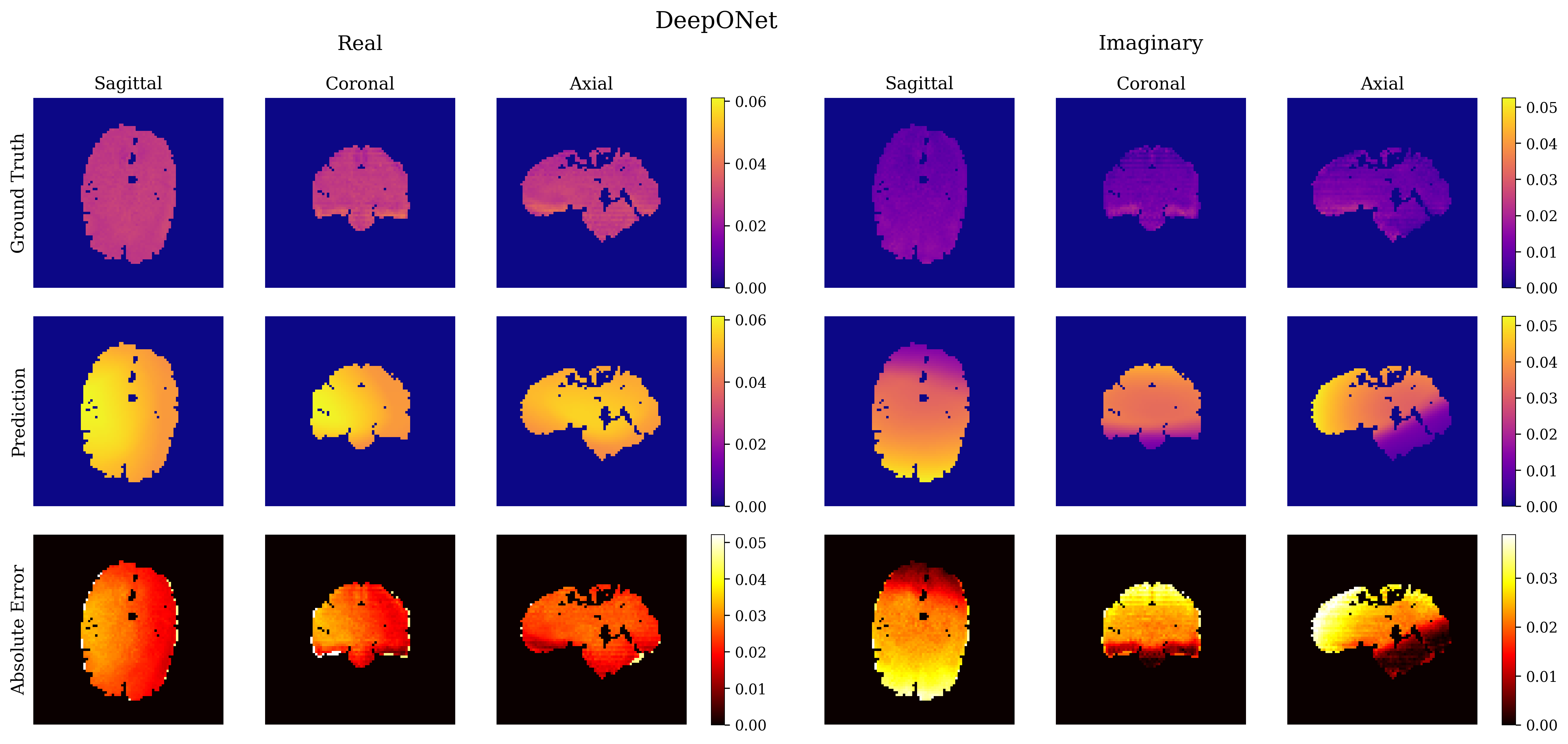}
  \caption{\DRS{DeepONet predictions and absolute error maps across sagittal, coronal, and axial planes for real (left) and imaginary (right) displacement fields. The predictions are visibly over-smoothed relative to ground truth, with loss of fine-scale spatial detail in both components. This smoothing bias produces higher absolute errors (peak $\sim$0.05 for real, $\sim$0.03 for imaginary) that are distributed broadly across the brain volume rather than localized to specific regions, reflecting the global nature of the branch-trunk representation.}}
  \label{fig:don_prediction}
\end{figure}

\DRS{DeepONet achieved the strongest performance on real displacement fields among all architectures, with the lowest MSE ($0.0039 \pm 0.0057$), lowest MAE ($0.0350 \pm 0.0314$), and highest accuracy ($0.9000 \pm 0.1465$). This result is notable given that visual inspection reveals the predictions to be over-smoothed relative to ground truth, with loss of fine-scale spatial detail (Figure~\ref{fig:don_prediction}). The apparent contradiction between strong quantitative metrics and visible smoothing can be attributed to the nature of the MSE loss: by producing spatially smooth predictions that closely track the mean displacement structure, DeepONet avoids the large localized errors (e.g., axis-aligned streaking in FNO, single-voxel outliers in F-FNO) that disproportionately inflate MSE in other architectures. However, this smoothing bias leads to underrepresentation of localized gradients and biomechanical heterogeneity, with errors distributed broadly across the brain volume rather than concentrated in specific regions.}

\DRS{For the imaginary displacement fields, performance was weaker, with an MSE of $0.0064 \pm 0.0106$ and accuracy of $0.8669 \pm 0.1739$, placing it behind MG-FNO ($0.0058 \pm 0.0126$) on this component. The imaginary fields in this sample exhibit finer spatial structure than the real component, increasing the reliance on resolving high-frequency content where the branch-trunk representation is less effective. This contrast highlights a dependence of DeepONet's effectiveness on the spectral content of the target field: it excels when the output is dominated by smooth, global variations but struggles when localized features carry significant energy.}

\DRS{From a computational standpoint, DeepONet was by far the most efficient architecture tested, achieving $3.83$ it/s with only $2.09$M trainable parameters and $4.11$ GB GPU memory (Table~\ref{tab:compute}). This represents a $5.9\times$ speedup over the next fastest model (FNO, $0.65$ it/s) and over two orders of magnitude fewer parameters than the baseline FNO ($1.42$B). These properties make DeepONet well suited for resource-constrained settings, provided that the loss of spatial fidelity in high-gradient regions can be addressed through architectural modifications such as multiscale trunk networks or attention-based fusion.}

\subsection{Influence of Demographic Features}

\DRS{To evaluate the influence of demographic features on prediction accuracy, we conducted an ablation study using the DeepONet architecture, progressively adding demographic covariates to a base model that includes the T1 image, spatial coordinates, excitation frequency, and vibration direction. Table~\ref{tab:ablation} reports the results.}

\begin{table}[h]
\centering
\DRS{
\caption{\DRS{Ablation study on demographic feature inclusion using DeepONet on real displacement fields. The base configuration includes T1 image, spatial coordinates, frequency, and direction. Demographic features are added progressively.}}
\label{tab:ablation}
\begin{tabular}{lll cc}
\toprule
\textbf{Configuration} & \textbf{Demographic Features} &
\textbf{MAE} & \textbf{Accuracy} \\
\midrule
No demographics & --- & 0.0456 & 0.8653 \\
+ Age & Age & 0.0397 & 0.8951 \\
+ Age, Sex & Age, Sex & 0.0379 & 0.8957 \\
Full model & Age, Sex, Brain Volume & 0.0350 & 0.9000 \\
\bottomrule
\end{tabular}
}
\end{table}

\DRS{The inclusion of age alone yields an approximately 3 percentage point gain in accuracy, consistent with the well-established influence of age on brain tissue viscoelastic properties \cite{sack2009impact}. Adding sex provides a modest further improvement, reflecting known sex-related differences in brain morphology and mechanical response. In the full model, we additionally include brain volume. This choice was motivated not primarily by predictive performance but by methodological considerations: brain volume is a well-known confounding variable in neuroimaging analyses, and conditioning the model on it helps ensure that predictions reflect tissue-level biomechanical properties rather than gross morphometric differences across subjects. This follows standard practice in neuroimaging studies where total intracranial or brain volume is included as a covariate to control for head-size effects \cite{buckner2004unified}.}

\section{Conclusions}
\label{sec:conclusions}

\DRS{In this work, we adapted and systematically compared neural operator architectures for modeling brain biomechanics under a multimodal input setting, integrating volumetric anatomical imaging with scalar demographic and acquisition parameters. Building on established fusion strategies from the operator learning literature, specifically field projection for FNO-based models \cite{wen2022u} and multi-branch encoding for DeepONet \cite{jin2022mionet}, we evaluated how these approaches perform in a biomedical prediction context characterized by inter-subject anatomical variability, mixed covariate types, and spatially heterogeneous tissue properties.}

\DRS{Through evaluation on 249 \textit{in vivo} MRE experiments, we found that no single architecture dominated across all criteria. DeepONet achieved the highest accuracy on real displacement fields (MSE $0.0039$, accuracy $0.90$) while requiring the fewest parameters (2.09M) and delivering the fastest inference (3.83 it/s). MG-FNO achieved the best performance on imaginary displacement fields (MSE $0.0058$,accuracy $0.88$) and the lowest GPU memory usage among FNO variants (7.12 GB), but was the slowest at inference (0.08 it/s) due to sequential patch processing. The baseline FNO and F-FNO offered intermediate accuracy but were hampered by axis-aligned spectral artifacts and single-voxel instabilities, respectively. These results demonstrate that multimodal fusion strategy and architectural choice jointly determine the trade-off between accuracy, spatial fidelity, and computational cost.}

Despite these advances, challenges remain. Models exhibited residual spectral bias, underpredicting localized, high frequency deformation modes most relevant to tissue-level injury risk. \textcolor{blue}{The study was also conducted on a relatively small subset of subjects with consistent data availability, which may limit generalizability across populations, imaging sites, and pathological conditions. Additionally, the dataset consists of healthy volunteers, and the models have not been validated on injured brains or clinical outcome measures. As such, the applicability of these models to clinical decision-making or injury risk assessment remains an open question. }

Future work should incorporate multiscale frequency learning, hybrid physics informed loss functions, and uncertainty quantification. Investigating attention-based fusion mechanisms may further enhance multimodal integration. \DRS{Ablation studies quantifying the individual contribution of demographic and acquisition features to prediction accuracy would help identify which inputs are most informative for displacement prediction.} Extending models to time resolved, nonlinear impact scenarios and validating on independent, multisite datasets will be essential for clinical translation. Overall, this study provides a step toward understanding how neural operators can be adapted to multimodal biomedical settings and highlights key challenges and opportunities for future research.

\section{Acknowledgments}
 The authors would like to acknowledge computing support provided by the Advanced Research Computing at Hopkins (ARCH) core facility at Johns Hopkins University and the Rockfish cluster. ARCH core facility (\url{rockfish.jhu.edu}) is supported by the National Science Foundation (NSF) grant number OAC1920103. The authors gratefully acknowledges the Brain Biomechanics Imaging Repository (\url{http://www.nitrc.org/projects/bbir}) for data which was supported supported under grants U01 NS112120 and R56 NS055951. The research efforts of DRS and SG are supported by National Science Foundation (NSF) under Grant No. 2436738. Any opinions, findings, conclusions, or recommendations expressed in this material are those of the author(s) and do not necessarily reflect the views of the funding organizations.

\section*{Author contributions}
\noindent Conceptualization: AA, DRS, SG  \\
Investigation: AA, DRS, SG \\
Visualization: AA, DRS, SG  \\
Supervision: SG \\
Writing - original draft: AA, DRS \\
Writing - review \& editing: AA, DRS, SG 

\section*{Data and code availability}
\noindent The data is publicly available on \url{http://www.nitrc.org/projects/bbir}. The neural operator codes that support the findings of this study are available on \url{https://github.com/Centrum-IntelliPhysics/Neural-Operator-for-Traumatic-Brain-Injury}.

\section*{Competing interests}
\noindent The authors declare no competing interest

\section*{Ethics Statement}

This study analyzed publicly available, de-identified data from the Brain Biomechanics Imaging Resources (BBIR) repository (\url{https://www.nitrc.org/projects/bbir}). All original data were collected under IRB-approved protocols at Washington University in St. Louis, the University of Delaware, and the National Institutes of Health, with written informed consent obtained from all participants (parental consent for minors). Data were de-identified in accordance with HIPAA regulations. As this constituted secondary analysis of existing, publicly available, anonymized data, no additional IRB approval was required at Johns Hopkins University. All procedures complied with relevant institutional guidelines and the GNU GPL v3.0 license. Original ethics approvals are documented in Bayly et al. (2021).


\bibliographystyle{unsrt}
\bibliography{references} 

\end{document}